\documentclass{article}
\usepackage{spconf,amsmath,graphicx}
\usepackage{xcolor}

\title{Automatic defect segmentation by unsupervised anomaly learning}
\name{Nati Ofir, Ran Yacobi, Omer Granoviter, Boris Levant and Ore Shtalrid}

\address{Applied Materials}
\begin{document}
	%
	\maketitle

	\begin{abstract}
	    This paper addresses the problem of defect segmentation in semiconductor manufacturing. The input of our segmentation is a scanning-electron-microscopy (SEM) image of the candidate defect region. We train a U-net shape network to segment defects using a dataset of clean background images. The samples of the training phase are produced automatically such that no manual labeling is required. To enrich the dataset of clean background samples, we apply defect implant augmentation. To that end, we apply a copy-and-paste of a random image patch in the clean specimen. To improve the robustness of the unlabeled data scenario, we train the features of the network with unsupervised learning methods and loss functions. Our experiments show that we succeed to segment real defects with high quality, even though our dataset contains no defect examples. Our approach performs accurately also on the problem of supervised and labeled defect segmentation.
	\end{abstract}
	
	
\section{Introduction} \label{sec:intro}
\begin{figure}[tbh]
	\centering
	\includegraphics[height=110px]{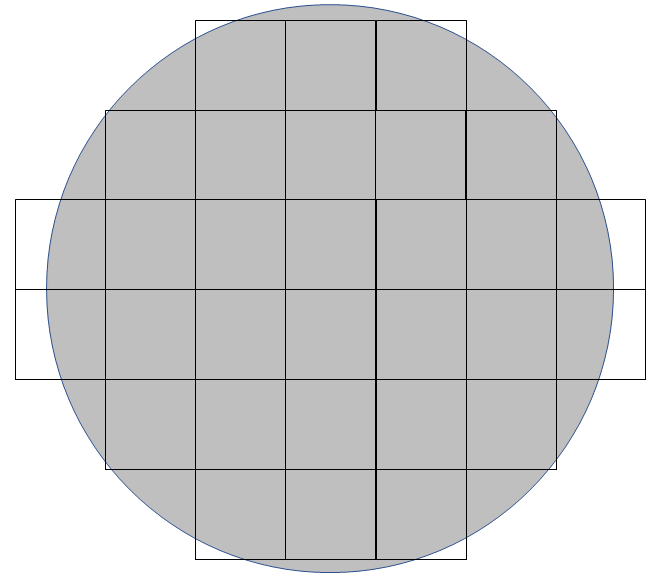}
	\caption{An illustration of a Silicon wafer and its corresponding dies. A specific die may be a defect candidate, while its corresponding location on another die is its reference clean image. The scanning electron microscopy (SEM) captures a small patch in a specific die such that the nano-meter semiconductor litography can be seen in high resolution.}
	\label{fig:wafer}
\end{figure}

This manuscript introduces a solution to the problem of automatic defect segmentation and detection. The uniqueness of our approach is that it learns from a non-defect image dataset, such that, no manual labeling or defect samples are required. The semiconductor manufacturing process applies nano-technology lithography to a bare wafer of Silicon. A single wafer contains several repetitive dies. For every point which is a candidate for a defect in the manufacturing process, we capture a SEM image, named a true defect image. Since the silicon slice contains many dies, for every such true defect we can find a non-defective reference image in another die. A true defect is equal to its reference image up to the defect appearance. See Figure \ref{fig:wafer} for an illustration of a silicon wafer with printed dies.

\begin{figure}[tbh]
	\centering
	\includegraphics[height=90px]{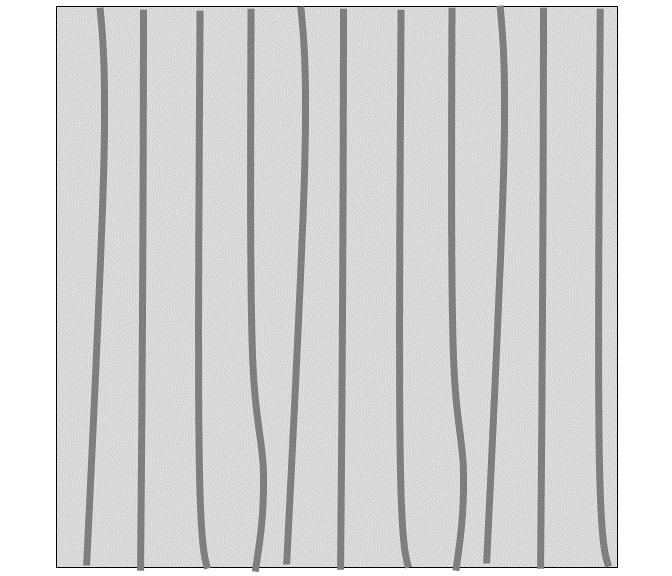}~
	\includegraphics[height=90px]{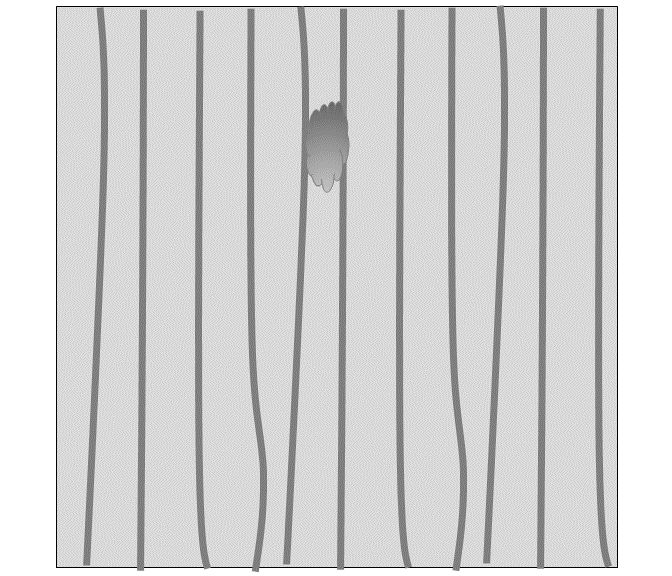}
	\caption{Left: Simulation Image of a semiconductor SEM image of a clean background pattern with lines. Right: The corresponding image with a defect between the lines. This defect type is called particle and it is relatively easy to detect and segment. Other defects may be low signal-to-noise-ratio defects that are more challenging. }
	\label{fig:refdef}
\end{figure}

A straightforward approach to detecting and segmenting defects is to compare the true defect to its reference. In the first step, applying registration between them by phase correlation for example \cite{foroosh2002extension}. Then detecting the maximum by a local threshold on the registered difference image, and detect the local maxima blobs by connected-component algorithm \cite{bailey2007single}. This traditional-classic approach segments the defects in a reference-defect scheme (Ref-Def). See Figure \ref{fig:refdef} for a simulation of a true-defect SEM image and its reference.

In our method, we tackle the problem of defect segmentation by a defect image, even without the help of a reference image, namely deep learning automatic defect review (DL-ADR). It means that we only capture the defect SEM image, and detect and segment its possible defects. Then, we apply a Deep-CNN to capture the defect in the single true defect image. In order for DL-ADR to be carried out by a supervised approach, manual labeling of defects is necessary. However, we introduce a method to apply DL-ADR without human intervention. Our approach takes advantage of the large number of clean reference images that we can capture in every Silicon wafer, and then uses this clean dataset of reference-background images.

In our solution, we implant simulated defects in the background semiconductor images, and train our CNN to segment and detect them. This process can be done with a supervised binary-cross-entropy (BCE) loss \cite{ho2019real}. To improve our robustness to the lack of labeled data, we use the semi-supervised approaches of dense contrastive-learning (Dense Sim-Clr) \cite{chen2020simple,wang2021dense} and teacher-student consistency \cite{chen2021semi}. Our experiments demonstrate that we succeed to segment real defects in SEM images, by training on defect-implanted reference images. Moreover, the contrastive regularization contributed to the quality of the learned CNN, as well as the teacher-student approach.

The paper outline is as follows. In Section \ref{sec:previous} we describe a literature review of the previous related works. In Section \ref{sec:semi} we introduce our CNN architecture for semi-supervised segmentation learning. Then, Section \ref{sec:bce} explains the variant of the BCE loss that we introduce. In addition, Section \ref{sec:dense} is on the dense contrastive learning loss we use to train our CNN. Section \ref{sec:aug} contains the augmentation description we apply to the reference images. The second alternative for semi-supervised learning, i.e. teacher-student consistency is described in Section \ref{sec:consistency}. Section \ref{sec:exp} contains the experiments, and we conclude our contribution in Section \ref{sec:conclusions}.

\section{Previous Work} \label{sec:previous}
The problem of image segmentation is a well-studied area with a plethora of related works. Early works relied on spatial and brightness properties of the image gray levels \cite{haralick1985image}. Advanced methods are based on classic machine learning (ML) algorithms like clustering by k-means \cite{dhanachandra2015image} or the expectation-minimization algorithm \cite{diplaros2007spatially}. A boundary marking by humans was introduced as a benchmarking segmentation method by Berkely-Segmentation-Dataset (BSDS) \cite{MartinFTM01}. Recent approaches aimed to develop solutions for maximizing the F-measure by training and testing on this labeled data \cite{xie2015holistically, arbelaez2014multiscale}. State-of-the-art methods are using deep-learning (DL), and encoder-decoder CNN to segment real images \cite{badrinarayanan2017segnet,chen2017deeplab}. The DL and classic approaches have each one their own strengths and limitations as described in \cite{ofir2021classic}. Our method performs image segmentation using DL, an encoder-decoder approach, and a semi-supervised loss. 

The area of defect detection is a fundamental task of computer vision and image processing. Classic approaches relied on Fourier-analysis \cite{chan2000fabric}. Other traditional approaches are based on thresholding an image by automatic derivations \cite{ng2006automatic}. A group of works is focused specifically on the problem of defect detection on the semiconductor wafer \cite{shankar2005defect}. Recent approaches address this problem as anomaly detection by generative-adversarial-networks (GAN's) \cite{deecke2018image}. Small defects may be very challenging and require detection at low SNR as described in \cite{ofir2019detection,ofir2021multi}. Our work applies detection as a post-processing phase to segmentation, and it is carried out by training a CNN on unlabeled augmented data. 

Semi-supervised learning is a challenging problem with recent related papers. Early methods addressed learning from unlabeled or weakly labeled data in the classic ML field \cite{zhu2009introduction}. Contrastive learning \cite{chen2020simple} was introduced to learn a representation from unlabeled data for image classification. Recent approaches use teacher-student architectures to constrain a consistency on the Siamese networks applied on the same input \cite{cai2019exploring,caron2021emerging,chen2021semi}. In our work, we improve the CNN training process by regularizing the loss with a dense-contrastive-learning approach for image segmentation \cite{wang2021dense}, and by the teacher-student consistency \cite{chen2021semi}.

\section{Semi-supervised Architecture} \label{sec:semi}
\begin{figure}[tbh]
	\centering
	\includegraphics[height=120px]{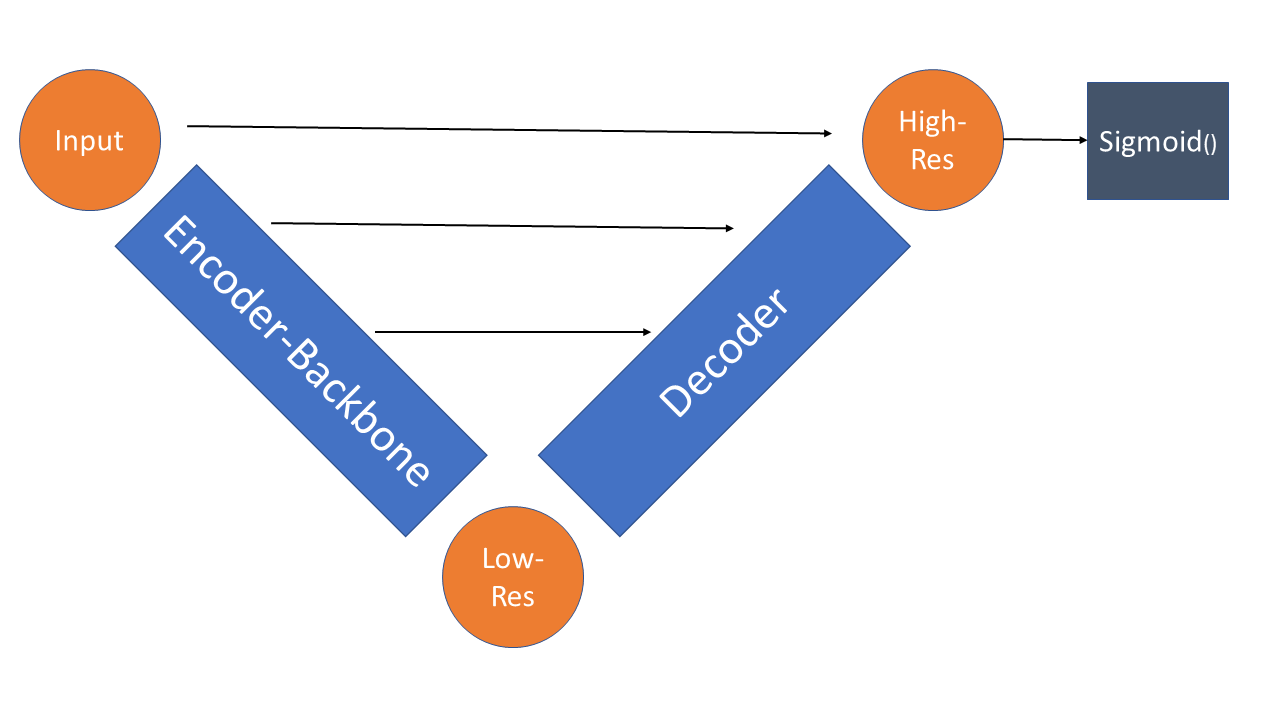}
	\caption{Schematic architecture of our U-net CNN. The weighted BCE segmentation loss is applied on the output of the Sigmoid activation function \cite{marreiros2008population}. The dense contrastive loss can be applied on the low-resolution or high-resolution features. Our solution that is described in the experiments Section \ref{sec:exp} is using the high-resolution features before the Sigmoid funtion. This multiscale architecture extracts informative features and segmentation maps for a given cadidate SEM image for a defect.}
	\label{fig:arc}
\end{figure}

For the purpose of defect segmentation, we introduce in this Section a CNN architecture with two heads, segmentation and unsupervised. To detect and segment defects in SEM images we use a U-net shape \cite{unet} network. The input to the network is an image with a candidate defect in the Silicon wafer. Then it goes through a forward pass in the encoder-backbone of the CNN. The output of this part is the low-resolution features, that can be used as an unsupervised head of the network. These features are forwarded through a decoder to infer the high-resolution features. We recommend using them as the unsupervised features, and the experiments reported in Section \ref{sec:exp} are evaluating this approach. We apply a projection head of 1X1 convolution before applying the unsupervised loss. The high-resolution features are projected to a probability map, and then activated by a Sigmoid function to get scores between zero to one, note that $Sigmoid(x) = \frac{e^x}{1+e^x}$. The output of the activations is the segmentation map that is the input for the supervised BCE loss. See Figure \ref{fig:arc} for illustration of our learning architecture.

Our network produces a segmentation map for every input image of a defect candidate. To transform it to detection maps we apply clustering post-processing such that the segmentation blob are clustered by ellipses. This post-processing is automatic, and it is used to compute our final metric. To conclude, our method produces segmentation maps together with detection locations.

\section{Weighted BCE Loss} \label{sec:bce}
We train our CNN segmentation head with a weighted Binary-Cross-Entropy (BCE) loss function. Denote by $p$ the segmentation output of the network, by $y$ the supervision labels, and by $w$ the weight that we compute according to the inputs. Then by iterating over the pixels, the weighted BCE loss formula is as follows:

\begin{equation} \label{eq:bce}
    WBCE(p,y) = \frac{1}{N}\sum_{i=1}^{N}w_iy_i\log{p_i}.
\end{equation}

The weights $w$ are computed such that we balance between the foreground and the background of the pixels, according to the precomputed labels.

\section{Contrastive Loss} \label{sec:dense}
To improve our learning of a feature representation, we add a dense contrastive learning (D-CLR) head to our CNN. We use the approach of dense contrastive learning as described in \cite{wang2021dense}. The input for this loss is $x$, the high-resolution features for example. The second input is an augmentation function $A$ which preserves the segmentation map. The function $A$ we use is Gaussian noise and a color-contrast jitter. The contrastive loss attracts an input to its augmentations, and distracts it from the negative samples as follows:

\begin{equation} \label{eq:clr}
    CLR(x,A) = -\frac{1}{N}\sum_{i=1}^{N}\log{\frac{\exp(sim(x_i,A(x_i))/\tau)}{\sum_j \exp(sim(x_i,A(x_j))/\tau)}}.
\end{equation}

In the dense approach, $x_i$ is a pixel feature vector, while $x_j$ is another pixel in other samples in the training batch. Therefore, the negative is different from our original sample. The temperature $\tau$ is calibrating the softmax function. The similarity function we use is cosine similarity, i.e. for two vectors $A,B$, the similarity is 
\begin{equation}
sim(A,B) =  cosine(A,B) = \frac{\sum A \cdot B}{||A||\cdot ||B||}.    
\end{equation}

\section{Augmentation and Enrichment} \label{sec:aug}

\begin{figure}[tbh]
	\centering
	\includegraphics[height=90px]{Reference.png}~
	\includegraphics[height=90px]{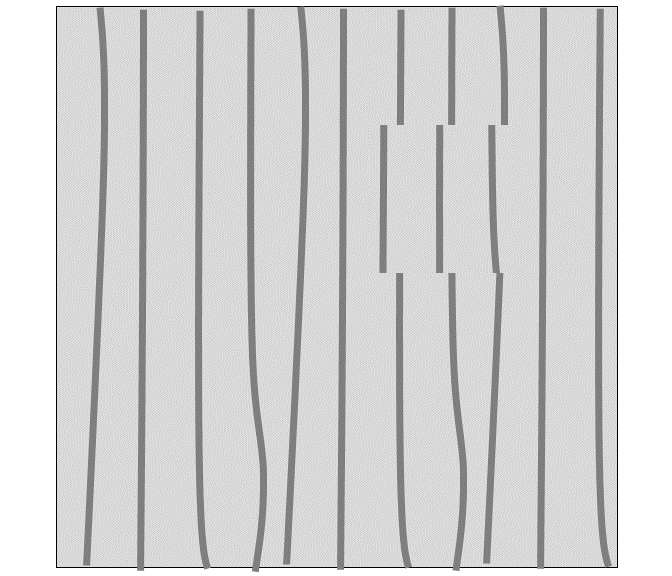}
	\caption{Left: Simulation image of a semiconductor SEM of a clean background pattern with lines. Right: The corresponding image with a copy paste augmentation, the copied area is marked as defected pixels. A copy paste augmentation is implanting an anomaly in the image that can be considered as a true-defect.}
	\label{fig:copypaste}
\end{figure}

To learn from unlabeled images, with the background only, we developed defect implant augmentations. The type of augmentation that we use is copy and paste. See Figure \ref{fig:copypaste} for an illustration of a copy-paste augmentation applied on a background image with no defects. The copy-paste is the copying of an image patch in the background image to another location. This copied patch is an anomaly in the image and can be seen as a defect in the reference image. By this augmentation, we implant defects in a clean image. The labels of this augmented image are updated corresponding to the patch pixel locations. Then it can be used for training with supervised BCE loss \eqref{eq:bce}, and by unsupervised dense contrastive loss \eqref{eq:clr}..

\section{Consistency Loss}\label{sec:consistency}
An alternative to the semi-supervised loss that was introduced in Section \ref{sec:dense} is to use a consistency loss instead as described in \cite{chen2021semi}. The idea is to train a network by an unsupervised loss that assumes a teacher-student consistency. It means that running two pseudo-Siamese networks, whose only difference is that they were initialized with different weights, on the same image input, should produce a consistent segmentation map. Therefore, given image input $I$, student network output $p_1(x) = CNN(I(x), \Theta_1)$ and teacher network $p_2(x) = CNN(I(x),\Theta_2)$. The consistency loss that can be used to improve our semisupervised training is as follows:
\begin{equation}
CL(p_1,p_2) = \sum_x p_1(x)\log(p_2(x))+\sum_x p_2(x)\log(p_1(x)),
\end{equation}
where $CL$ stands for consistency loss.

\section{Experiments}\label{sec:exp}

\begin{figure}[tbh]
	\centering
	\includegraphics[height=100px]{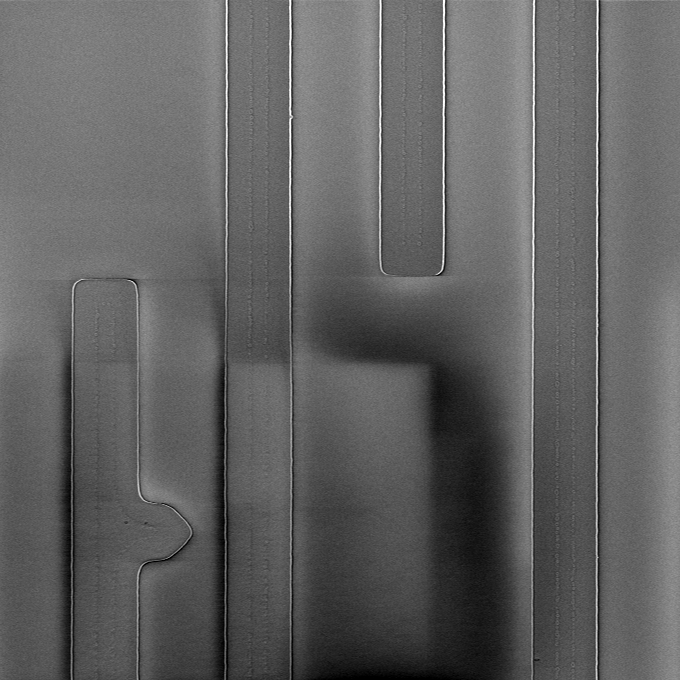}~
	\includegraphics[height=100px]{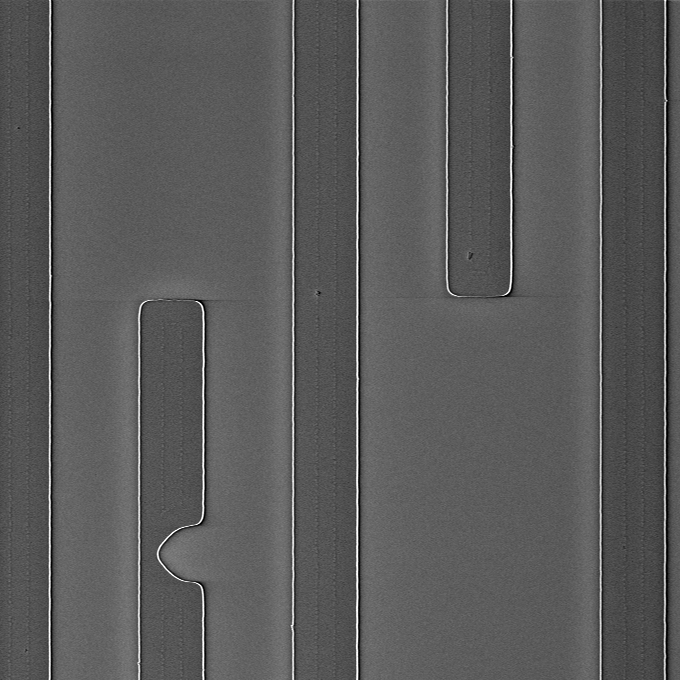}
	\caption{Simulation of true-defects in SEM images. These specific simulations contain defects which are relatively easy to detect and segment.}
	\label{fig:sim}
\end{figure}

We tested our approach on a real dataset of defect images in the semiconductor manufacturing process. The dataset contains approximately 4000 real samples of background-reference images. The test set contains approximately 1000 true-defect images of semiconductors SEM images. Figure \ref{fig:sim} shows samples of patterns of SEM images. To compute a metric for our performance we evaluate our method as a detection algorithm. We classify the detections, on top of our segmentation to hit, miss, false alarm, and filtered. According to these classifications, we compute the precision and recall of our CNN.

\begin{figure}[tbh]
	\centering
	\includegraphics[height=120px]{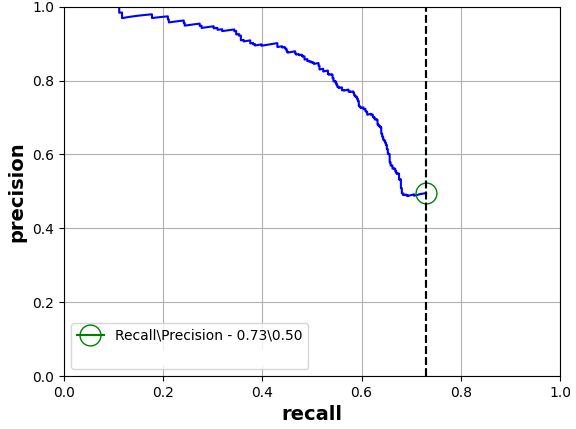}
	\caption{Precision-recall curve of our approach using dense contrastive learning and a weighted BCE loss. The point with maximum recall is marked. It can be seen that we achieve high performance at every working point.}
	\label{fig:pr}
\end{figure}

Table \ref{table:f} describes the F-measure of our results, which is the harmonic mean of precision and recall, $F = \frac{2\cdot P\cdot R}{P+R}$. It can be seen that our approach achieves high-quality results, with respect to the fact that no human labeling and true-defect images are involved in the training phase. Using the unsupervised loss is superior to using BCE-only loss. Note that previous approaches usually require reference images and therefore we do not add them to this table. Figure \ref{fig:pr} shows the precision-recall graph of our solution, it can be seen that with different thresholding we can achieve a recall of 0.73. The graphs show that our CNN produces meaningful full segmentation maps at every working point. See Table \ref{table:supervised} for our performance when working in a regular mode while training the CNN with human labeling. It can be seen that our approach can be transformed into supervised anomaly detection with high quality.

\begin{table}[tbh]
	\centering%
	
	\begin{tabular}{|l|c|c|c|}
		\hline
		Unsupervised Algorithm & F-Measure & Precision & Recall \\ \hline\hline
		D-CLR & \textbf{0.65} & 0.65 & 0.64 \\ \hline
		W-BCE & 0.62 & 0.7 & 0.55\\ \hline
	\end{tabular}%
	\caption{Table of F-Measure of defect detection accorss the different methods of our unsupervised approaches: D-CLR (with W-BCE) and W-BCE only. Each loss is tested with Copy and Paste augmentation. Unsupervised D-CLR added to W-BCE is improving the overall F-measure.}
	\label{table:f}
\end{table}

\begin{table}[tbh]
	\centering%
	
	\begin{tabular}{|l|c|c|c|}
		\hline
		Algorithm & F-Measure & Precision & Recall \\ \hline\hline
		Teacher-Student-Ref & \textbf{0.87} & 0.83 & 0.92 \\ \hline
		D-CLR-Ref & 0.86 & 0.82 & 0.89 \\ \hline
		W-BCE-Ref & 0.85 & 0.84 & 0.86\\ \hline
		Teacher-Student & \textbf{0.8} & 0.75 & 0.85 \\ \hline
		D-CLR & 0.74 & 0.69 & 0.78 \\ \hline
		W-BCE & 0.82 & 0.76 & 0.87\\ \hline
		Classic-Ref & 0.73 & 0.59 & 0.92\\ \hline
	\end{tabular}%
	\caption{Table of F-Measure of defect detection by fully supervised learning using human labeling for comparison with our unsupervised approach. The compared methods are Teacher-Student consistency, Dense-Contrastive-Learning (D-CLR), weighted Binary-Cross-Entropy (W-BCE) and classic defect segmentation using reference image. Top results are achieved when working in Ref-def mode, such that the input for the CNN is the defect image together with its corresponding clean reference image. Ref-def results are better than working in a defect-only DL-ADR.}
	\label{table:supervised}
\end{table}

The experiments of this Section demonstrate that our approach for unsupervised anomaly learning produces meaningful segmentation maps and detection for input images of candidate defects in semiconductor manufacturing. In addition, the semi-supervised loss is contributing to the quality of the results with respect to applying supervised training only in augmentation locations.

\section{Conclusions} \label{sec:conclusions}

This paper introduced an end-to-end solution to defect segmentation and detection learning using a clean dataset of background images. We presented a CNN U-net architecture with multi-head-trained by a semi-supervised loss function. For the supervised loss, we introduced a weighted BCE approach. For the unsupervised head, we trained using a dense contrastive approach with segmentation preserving augmentations, and teacher-student consistency. To enrich our background dataset with defects, we used copy-paste augmentations. As a whole study, this manuscript describes a solution to defect segmentation in the semiconductor manufacturing process with no manual labeling involved.
	{\small
		\bibliographystyle{ieee}
		\bibliography{egbib}
	}
	
\end{document}